\documentclass{article}

\PassOptionsToPackage{numbers, compress}{natbib}

\usepackage[final]{nips_2018}

\usepackage[utf8]{inputenc} 
\usepackage[T1]{fontenc}    
\usepackage{hyperref}       
\usepackage{url}            
\usepackage{booktabs}       
\usepackage{amsfonts}       
\usepackage{nicefrac}       
\usepackage{microtype}      
\usepackage{graphicx}

\title{General-to-Detailed GAN for \\ Infrequent Class Medical Images}

\author{
  Tatsuki Koga\thanks{Medical Sciences Innovation Hub Program, RIKEN
  and Center for Advanced Intelligence Project, RIKEN} \\
  \texttt{tatsuki.koga@riken.jp}
  \And
  Naoki Nonaka\thanks{Medical Sciences Innovation Hub Program, RIKEN} \\
  \texttt{naoki.nonaka@riken.jp}
  \AND
  Jun Sakuma\thanks{Center for Advanced Intelligence Project, RIKEN and Department of Computer Science, University of Tsukuba} \\
  \texttt{jun.sakuma@riken.jp}
  \And
  Jun Seita\thanks{Medical Sciences Innovation Hub Program, RIKEN
  and Center for Advanced Intelligence Project, RIKEN} \\
  \texttt{jun.seita@riken.jp}
}

\begin{document}

\maketitle

\begin{abstract}
  Deep learning has significant potential for medical imaging. However, since the incident rate of each disease varies widely, the frequency of classes in a medical image dataset is imbalanced, leading to poor accuracy for such infrequent classes. One possible solution is data augmentation of infrequent classes using synthesized images created by Generative Adversarial Networks (GANs), but conventional GANs also require certain amount of images to learn. To overcome this limitation, here we propose General-to-detailed GAN (GDGAN), serially connected two GANs, one for general labels and the other for detailed labels. GDGAN produced diverse medical images, and the network trained with an augmented dataset outperformed other networks using existing methods with respect to Area-Under-Curve (AUC) of Receiver Operating Characteristic (ROC) curve.
\end{abstract}

\section{Introduction}

Deep convolutional networks have made a breakthrough in the field of image classification. In the medical imaging field, many applications have been developed utilizing open datasets such as Chest X-ray dataset from National Institutes of Health (NIH) \cite{wang2017chestx}. \\
The typical model for medical imaging diagnosis takes images and auxiliary data as inputs, and outputs whether the data indicates diseases. While the task itself does not so much differ from the task in other fields, the model may fail to learn features well due to medical image specific issues. One main issue causing less accurate models is imbalance of data availability. The number of data without diseases is much larger than those with diseases. In addition, while some diseases commonly appear in the dataset, others appear very rarely. In general, it is harder for models to learn well on the relatively infrequent classes due to the amount of the information given to it. To overcome the imbalance, oversampling and undersampling are typical techniques \cite{ganganwar2012overview}, and derivative algorithms have been proposed \cite{han2005borderline}. The most straightforward way to oversample is to randomly duplicate some data with minor classes and add this to a train dataset. The way to undersample is to remove some data with major classes from the dataset. In general, oversampling leads to overfitting and undersampling leads to loss of the information. \\
Generative Adversarial Networks (GAN), originally proposed by Goodfellow et al \cite{goodfellow2014generative}, generate realistic synthetic samples from high-dimensional distributions. Some GANs, such as auxiliary classifier GAN (ACGAN) \cite{odena2016conditional}, can generate images with a given class. However, conventional GANs also require certain amount of train dataset. \\
In order to deal with the imbalance of the medical image, we propose in this paper a General-to-detailed GAN (GDGAN) to generate medical images with infrequent classes. GDGAN contains two stages of GANs. The first GAN generates medical images with general classes, such as gender, and the second GAN adds disease specific detailed classes to the images generated by the first GAN. \\

\section{Related Works}

Dataset augmentation for non-medical datasets utilizing GANs has been reported in 2017 \cite{antoniou2017data}. Although the datasets used did not have class imbalance, the augmentation generally improved the performance of deep convolutional networks. \\
The synthesis has been done for relatively small medical datasets as well \cite{guibas2017synthetic} \cite{frid2018gan}. They aimed to increase the total number of data in the dataset. Xin, Ekta and Paul tackled the imbalance of medical images by using GAN for a future extraction \cite{yi2018unsupervised}. However, they tested the effect of GAN only by support vector machine. \\

\section{Method}

\begin{figure}[htb]
  \centering
  \includegraphics[width=\linewidth]{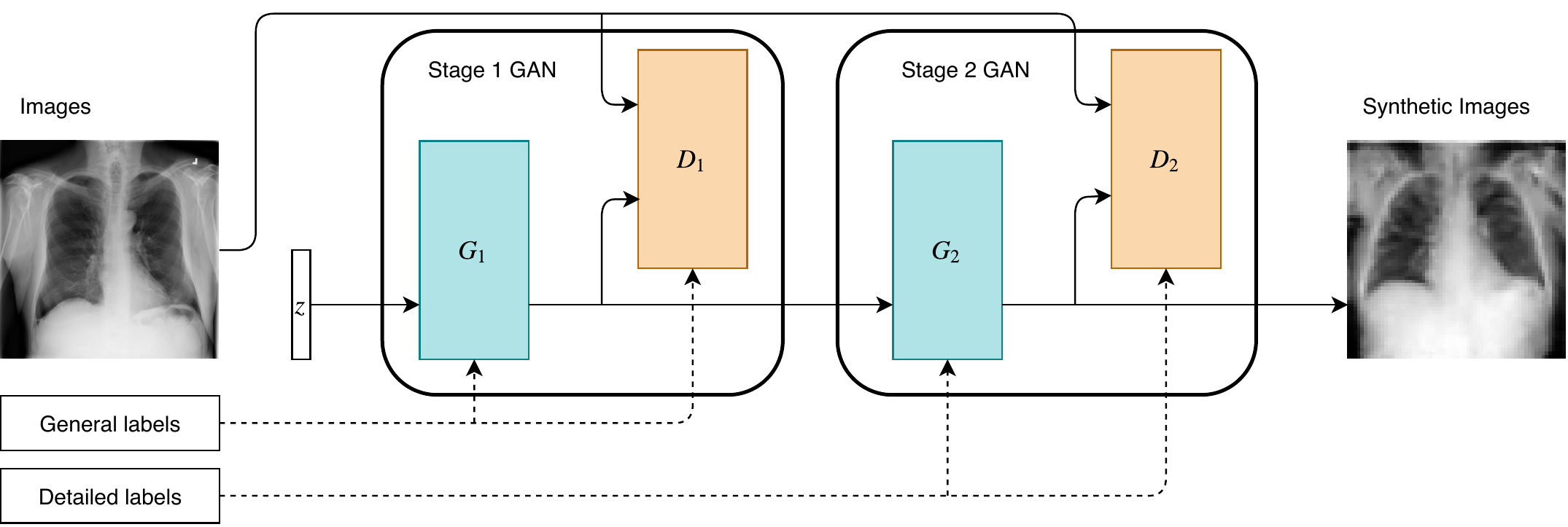}
  \caption{Flowchart of GDGAN.}
  \label{fig:flowchart}
\end{figure}

In general, medical image datasets contain both labels for general features, e.g., “male”, “female”, and labels with disease specific detailed features. To use this nature effectively, we designed a GAN named General-to-detailed GAN (GDGAN). GDGAN consists of two serially connected GANs. The first GAN takes a uniform noise distribution Z and general labels, e.g., gender, as inputs, then generates diverse images. The second GAN takes images generated by the first GAN, and detailed labels such as disease classes as inputs, and generates the final images (Figure~\ref{fig:flowchart}). \\
The loss function of the first GAN is partly taken from Wasserstein GAN (WGAN) with gradient penalty \cite{arjovsky2017wasserstein} \cite{gulrajani2017improved}, and also has the log-likelihood of the correct basic classes. The second GAN also has the loss of WGAN with gradient penalty and, in addition to that, it has log-likelihoods of the correct disease classes and basic classes (using the discriminator of the first GAN), and mean squared error loss between inputs and outputs. \\
The Chest X-ray dataset from NIH was used for the experiment. It contains 112,120 images in total and each image has one or more labels: age, gender, view position and 14 disease-specific phenotypes. We split the dataset into a train dataset (about 70\%), a validation dataset (about 10\%) and a test dataset (about 20\%). Two “general” labels, gender and view position, were fed to the first GAN, and “detailed” labels, such as Cardiomegaly (2,772 images, out 2.5\% of total images), were fed to the second GAN\@. Images were resized to a size of $64 \times 64$. Each GAN in GDGAN was trained separately. The first GAN was trained initially, then training of the second GAN followed. For each training, we used Adam \cite{kingma2014adam} as an optimizer with a learning rate of 0.0002. To represent single stage GAN, ACGAN was trained, too. \\
The diversity of images synthesized by ACGAN and GDGAN were compared to the diversity of input images using the inception score \cite{salimans2016improved}, which uses an Inception v3 Network \cite{szegedy2016rethinking} pre-trained on ImageNet \cite{deng2009imagenet}. \\
Following the original proposal \cite{gulrajani2017improved}, 50,000 images were randomly selected from Chest X-ray dataset, or generated by ACGAN and GDGAN, then split into 10 batches (5,000 images each), and the inception score was computed for each batch. Welch’s t-test was performed to determine whether the scores were significantly different. \\
Next, the VGGNet-19 \cite{simonyan2014very} model was built for a disease classifier, with a little modification to deal with grayscale input images. The network was trained with Chest X-ray dataset augmented in five ways: no augmentation ($N$=78,468$\pm$0, $N_{Cardiomegaly}$=1,937$\pm$13 for training), undersampled ($N$=64,276$\pm$243, $N_{Cardiomegaly}$=1,937$\pm$13), oversampled ($N$=100,716$\pm$119, $N_{Cardiomegaly}$=14,103$\pm$62), supplied by ACGAN ($N$=100,716$\pm$119, $N_{Cardiomegaly}$=14,103$\pm$62) and supplied by GDGAN ($N$=100,716$\pm$119, $N_{Cardiomegaly}$=14,103$\pm$62). For each method, we calculated an Area-Under-Curve (AUC) of ROC. We performed this experiment three times with different splits of the dataset.

\section{Results}

\begin{figure}[htb]
  \centering
  \includegraphics[width=0.25\linewidth]{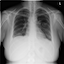}
  \includegraphics[width=0.25\linewidth]{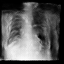}
  \includegraphics[width=0.25\linewidth]{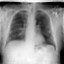}
  \caption{Randomly selected Chest X-ray dataset image, resized to a size of $64 \times 64$ (left) and synthetic image by ACGAN (center) and GDGAN (right).}
  \label{fig:images}
\end{figure}

\subsection{Inception score}

\begin{table}[htb]
  \caption{Mean inception scores and their standard deviation (SD) of Chest X-ray dataset and synthetic images by ACGAN and GDGAN (N=10).}
  \label{table:inception}
  \centering
  \begin{tabular}{lccc}
    \toprule
    & Chest X-ray dataset & Images by ACGAN & Images by GDGAN \\
    \midrule
    Mean score & 2.1258 & 2.7716 & 2.3220 \\
    SD & 0.0103 & 0.0172 & 0.0148 \\
    \bottomrule
  \end{tabular}
\end{table}

Table~\ref{table:inception} shows inception scores using 50,000 images of Chest X-ray dataset and synthetic datasets, by ACGAN and by GDGAN (examples shown in Figure~\ref{fig:images}). As the images were grayscale and only composed of X-ray images, both scores were not so high as other datasets such as CIFAR-10 \cite{barratt2018note}. However, the score of synthetic images by GDGAN was higher than that of Chest X-ray dataset (p=4.3748e-16). The score of synthetic images by ACGAN was the highest (p=4.6421e-22 for Chest X-ray dataset and p=9.4031e-22 for synthetic images by GDGAN). \\

\subsection{AUCs of ROC}

\begin{figure}[htb]
  \centering
  \includegraphics[trim=50 0 50 0, clip, height=0.27\linewidth]{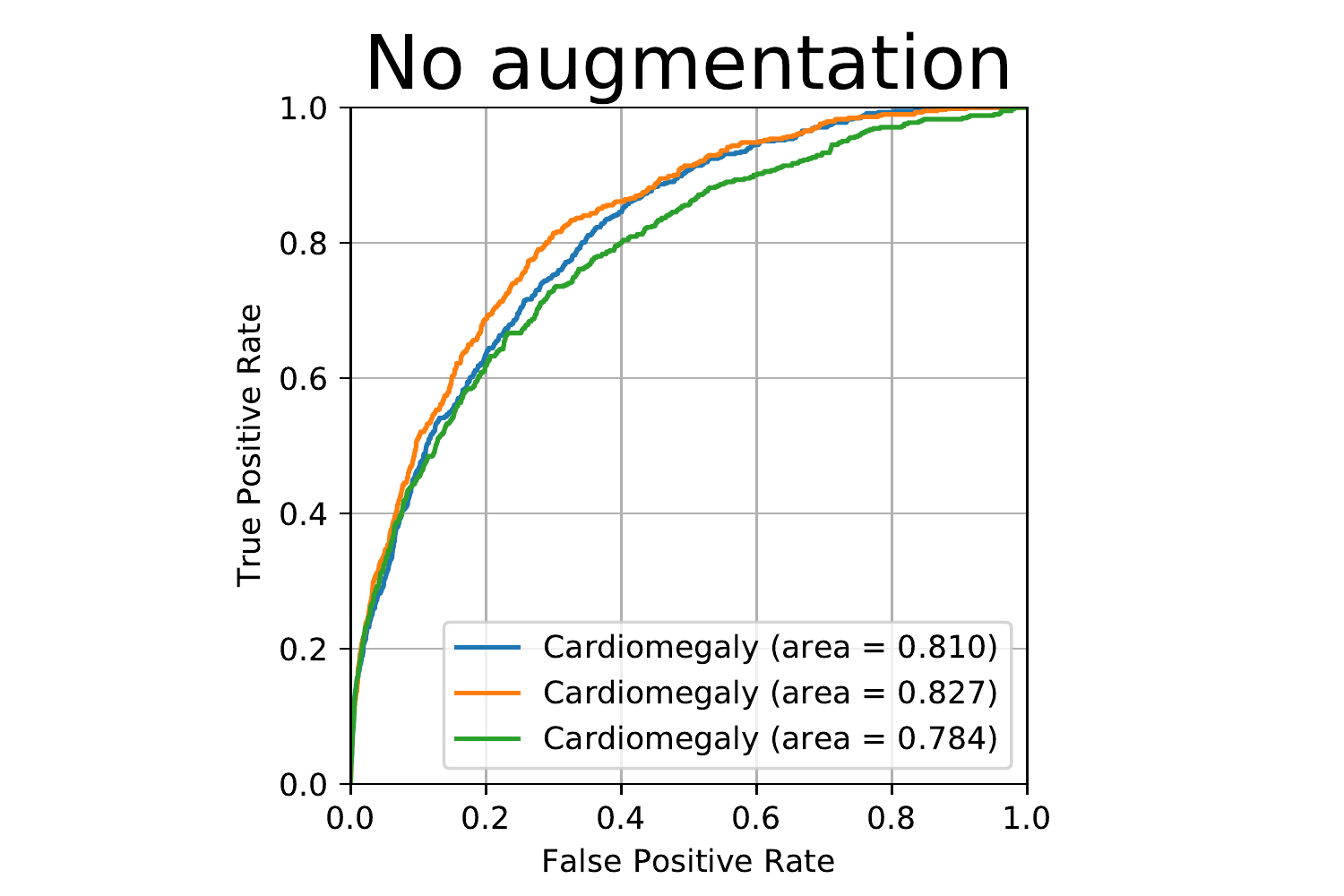}
  \includegraphics[trim=50 0 50 0, clip, height=0.27\linewidth]{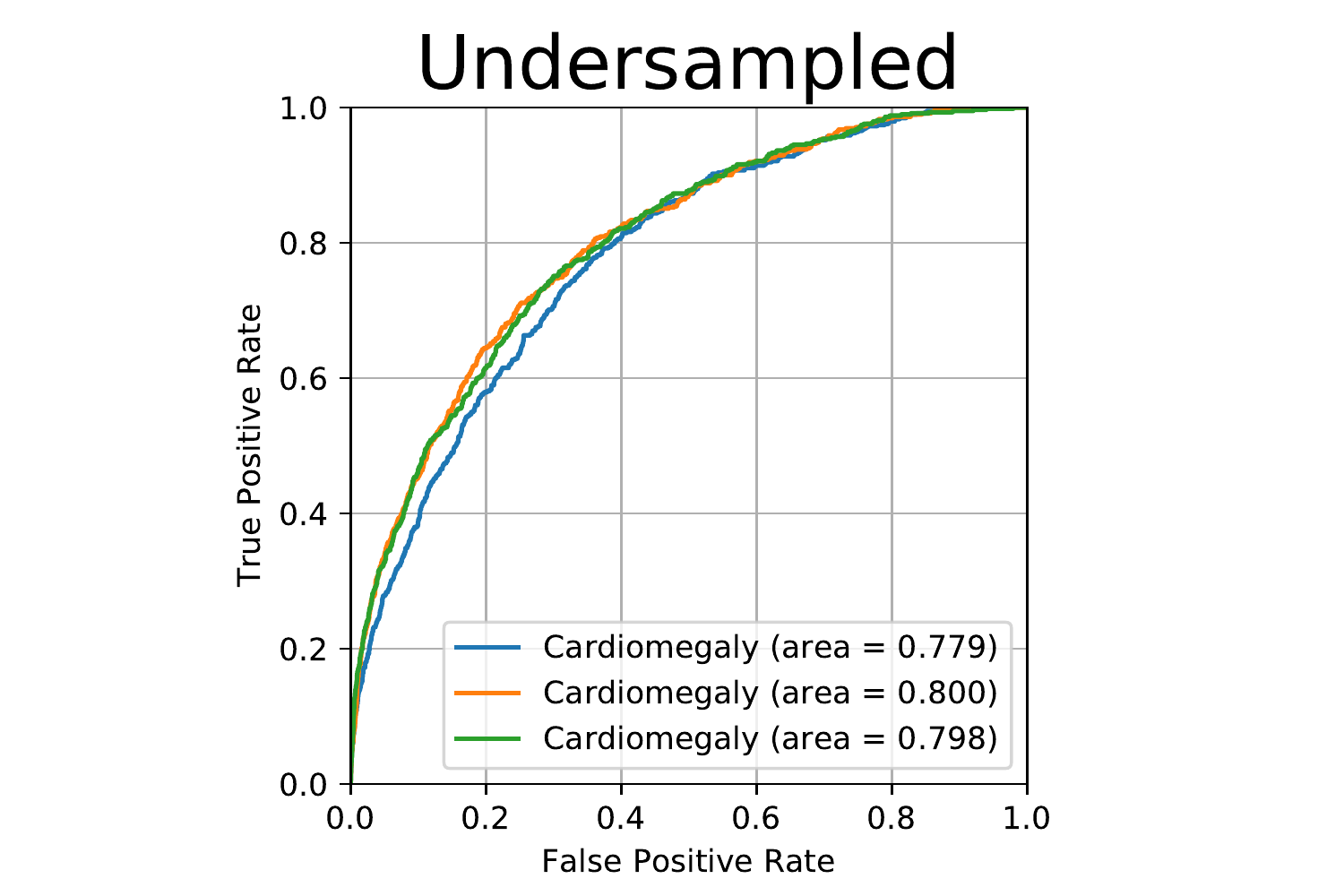}
  \includegraphics[trim=50 0 50 0, clip, height=0.27\linewidth]{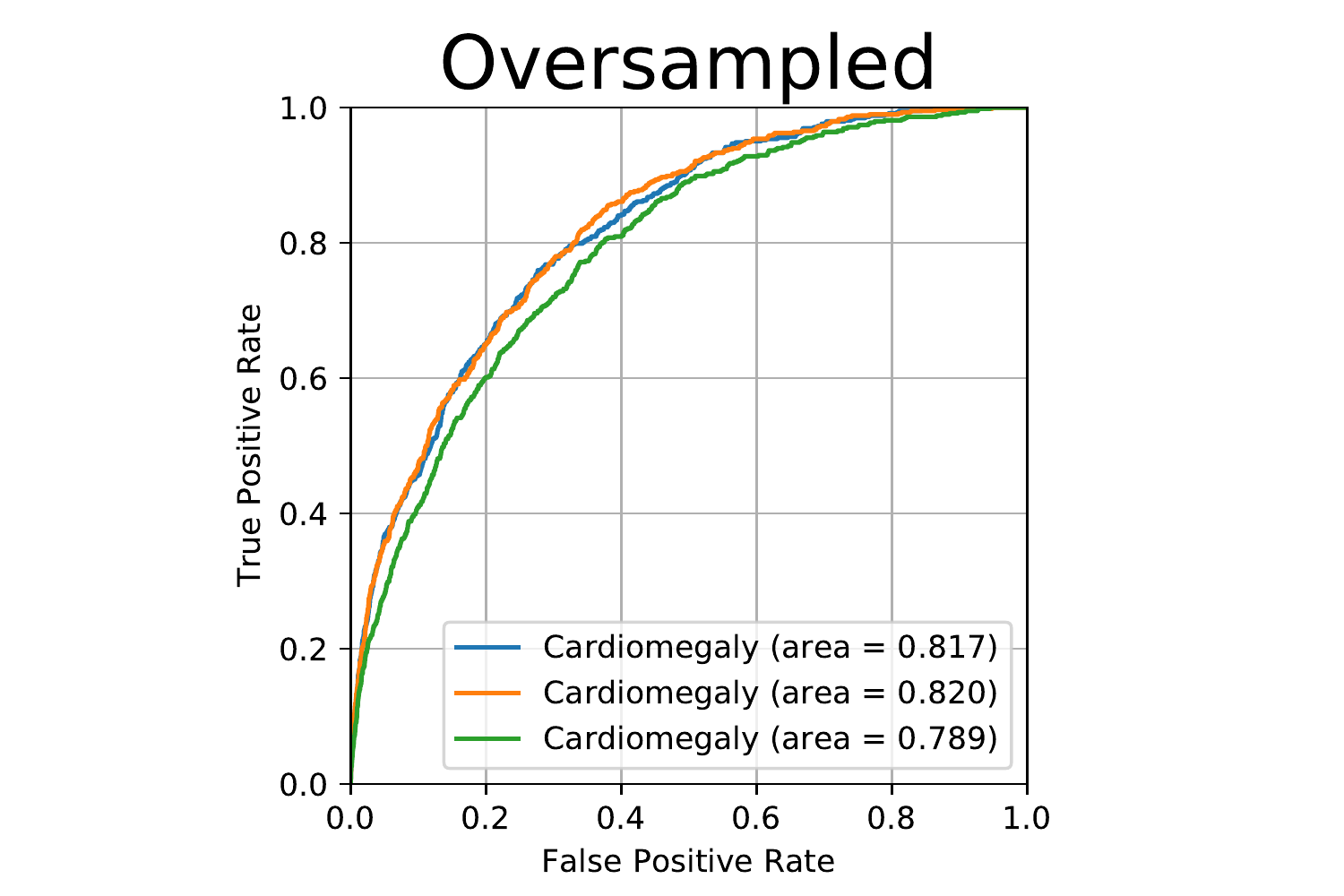}
  \includegraphics[trim=50 0 50 0, clip, height=0.27\linewidth]{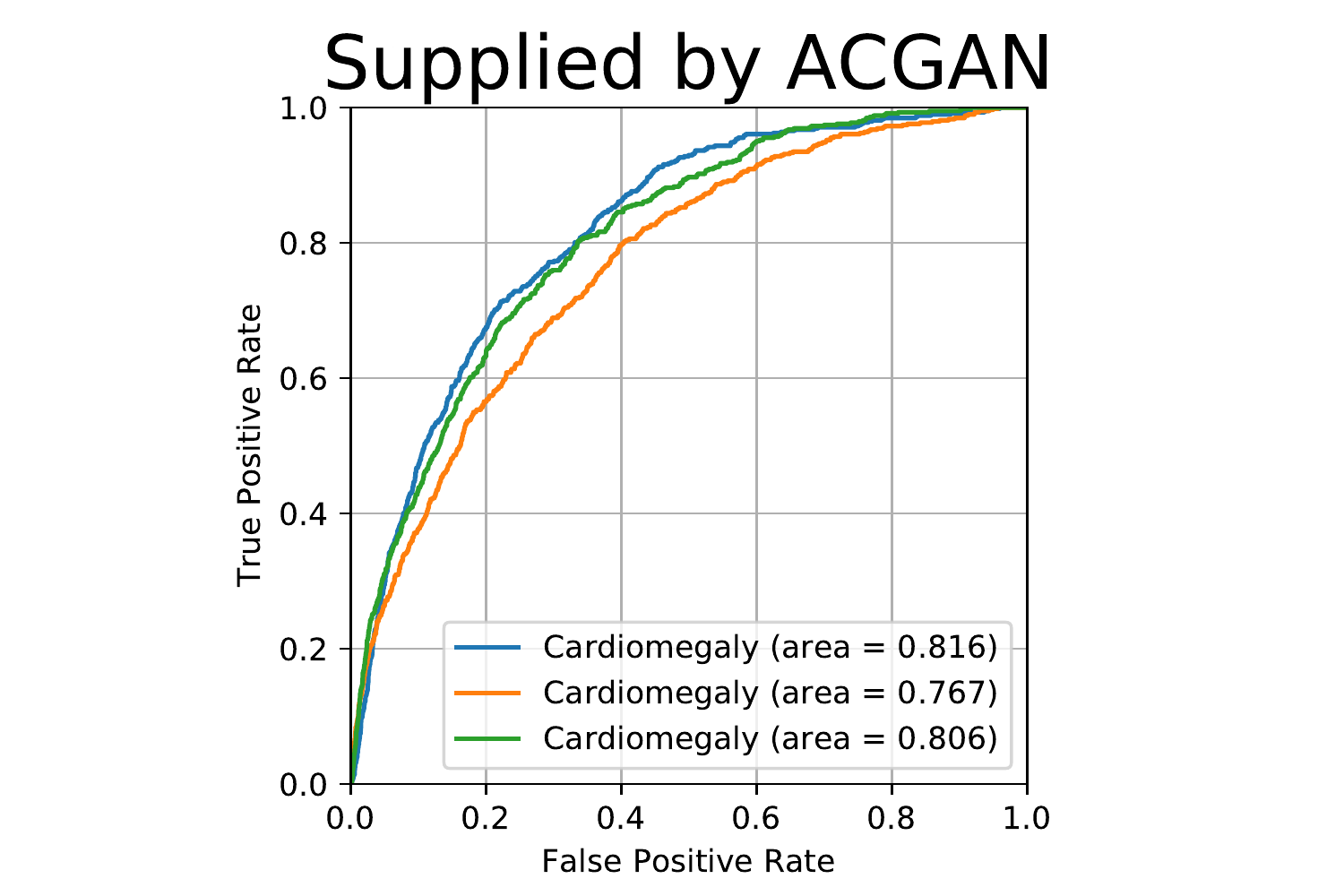}
  \includegraphics[trim=50 0 50 0, clip, height=0.27\linewidth]{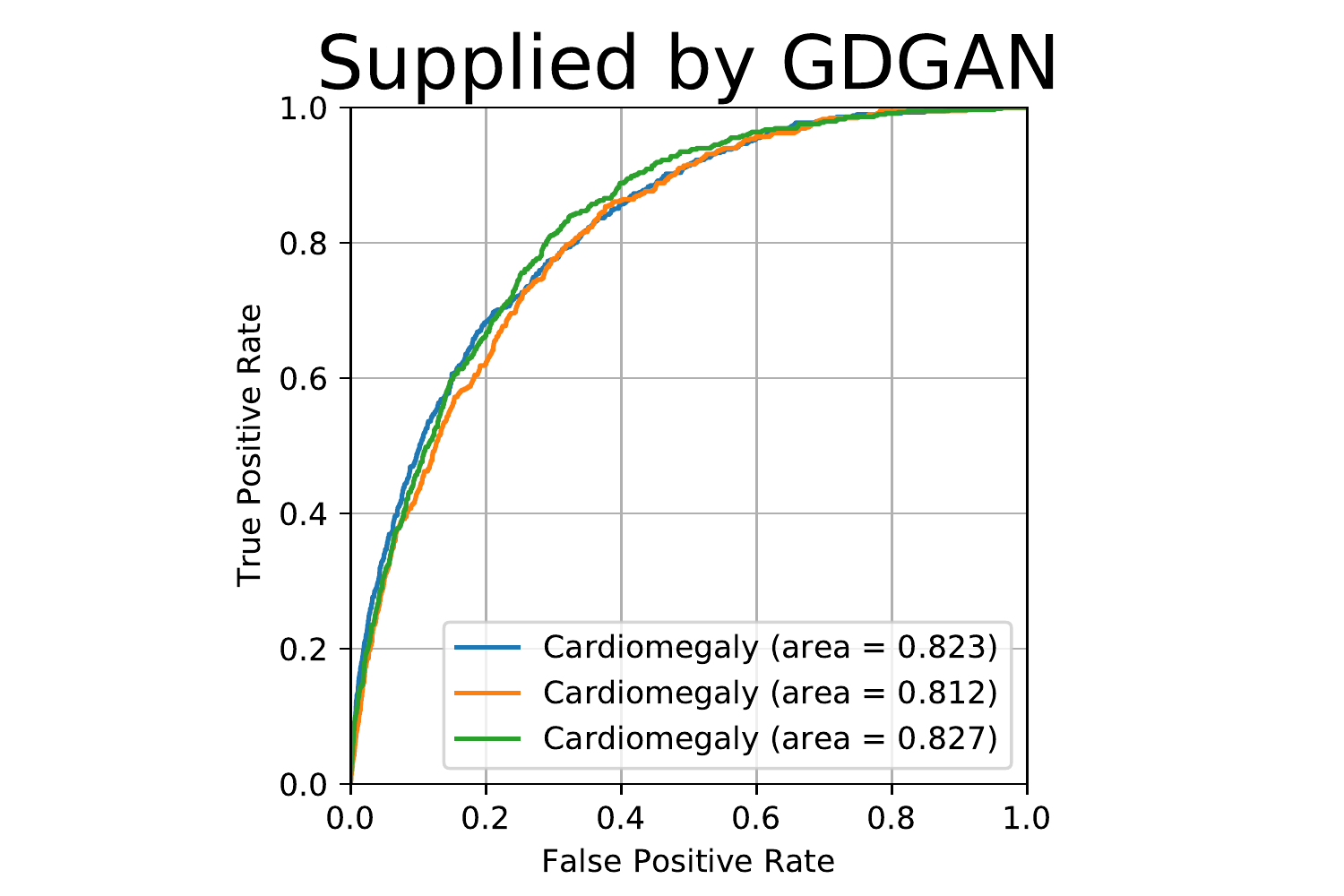}
  \caption{VGGNet-19 classification ROC curves of Cardiomegaly trained with Chest X-ray dataset augmented in different ways. Each curve represents the result of one of the experiments (blue: 1st, orange: 2nd, green: 3rd).}
  \label{fig:rocs}
\end{figure}

\begin{table}[htb]
  \caption{Comparison of total number of images for training ($N_{total}$), total number of images with Cardiomegaly ($N_{Cardiomegaly}$), VGGNet-19 classification mean AUC of ROC of Cardiomegaly and their standard deviation (SD) among different augmentation methods (N=3).}
  \label{table:aucs}
  \centering
  \begin{tabular}{lccccc}
    \toprule
    & Original & \multicolumn{4}{c}{Augmentation} \\
    \cline{3-6}
    & Chest X-ray & Undersample & Oversample & ACGAN & GDGAN \\ 
    \midrule
    $N_{total}$ & 78,468$\pm$0 & 64,276$\pm$243 & 100,716$\pm$119 & 100,716$\pm$119 & 100,716$\pm$119  \\
    $N_{Cardiomegaly}$ & 1,937$\pm$13 & 1,937$\pm$13 & 14,103$\pm$62 & 14,103$\pm$62 & 14,103$\pm$62 \\
    Mean AUC & 0.8074 & 0.7919 & 0.8084 & 0.7963 & {\bf 0.8207} \\
    SD & 0.0177 & 0.0095 & 0.0140 & 0.0212 & 0.0066 \\
    \bottomrule
  \end{tabular}
\end{table}

Figure~\ref{fig:rocs} demonstrates VGGNet-19 classification ROC curves of Cardiomegaly with different augmentation methods (no augmentation, undersampling, oversampling, supplying synthetic images by ACGAN and supplying synthetic images by GDGAN) to the Chest X-ray dataset. 
Table~\ref{table:aucs} shows the total number of images used in training, the total number of images with Cardiomegaly and the AUC of ROC (mean and standard deviation) corresponding to Figure~\ref{fig:rocs}. 
In comparison with the model trained without data augmentations, the models trained with undersampled Chest X-ray dataset and the dataset with synthesis images by ACGAN did not outperform. In contrast, the models trained with oversampled Chest X-ray and the dataset with synthesis images by GDGAN outperformed. And the model trained with Chest X-ray dataset with synthesis images by GDGAN scored the best AUC for Cardiomegaly. \\

\section{Discussion}

The comparison of inception scores revealed that GDGAN successfully generated infrequent class images as diverse as the original Chest X-ray dataset. Furthermore, supplying infrequent class images by GDGAN improved the performance of VGGNet-19 deep convolutional network quantified by AUCs of ROC\@.
One note is that we only tested relatively low resolution of images ($64 \times 64$) and one classification network. When generating and/or supplying higher resolution images by GDGAN and/or using other deep convolutional networks for evaluation, performance may alter.
Taken together, GDGAN, combination of a GAN for general classes and a GAN for detailed classes may open up a new strategy for deep learning with infrequent classes.

\subsubsection*{Acknowledgments}

We gratefully acknowledge the support of NVIDIA Corporation for the donation of the GPU used for this research. \\

\newpage
\bibliography{ref}
\bibliographystyle{unsrt}

\end{document}